\documentclass[10pt,twocolumn,letterpaper]{article}

\usepackage{cvpr}              

\usepackage{graphicx}
\usepackage{amsmath}
\usepackage{amssymb}
\usepackage{booktabs}

\usepackage{amsmath,amsfonts,bm}

\def\eqref#1{equation~\ref{#1}}

\def\1{\bm{1}}

\def\mF{{\bm{F}}}

\DeclareMathAlphabet{\mathsfit}{\encodingdefault}{\sfdefault}{m}{sl}
\SetMathAlphabet{\mathsfit}{bold}{\encodingdefault}{\sfdefault}{bx}{n}

\usepackage{multirow}

\usepackage[pagebackref,breaklinks,colorlinks]{hyperref}

\usepackage[capitalize]{cleveref}
\crefname{section}{Sec.}{Secs.}
\Crefname{section}{Section}{Sections}
\Crefname{table}{Table}{Tables}
\crefname{table}{Tab.}{Tabs.}

\def\cvprPaperID{*****} 
\def\confName{CVPR}
\def\confYear{2023}

\begin{document}

\title{A Study on Differentiable Logic and LLMs for EPIC-KITCHENS-100 Unsupervised Domain Adaptation Challenge for Action Recognition 2023}

\author{Yi Cheng$^{1,2}$, Ziwei Xu$^{2}$, Fen Fang$^{1}$, Dongyun Lin$^{1}$, Hehe Fan$^{3}$, Yongkang Wong$^{2}$ \\
Ying Sun$^{1}$, Mohan Kankanhalli$^{2}$ \\
$^{1}$Institute for Infocomm Research, Agency for Science, Technology and Research, Singapore \\
$^{2}$National University of Singapore\\
$^{3}$Zhejiang University\\
{\tt\small \{chengyi,ziwei-xu,mohan\}@comp.nus.edu.sg, yongkang.wong@nus.edu.sg,}\\
{\tt\small \{fang\_fen,lin\_dongyun,suny\}@i2r.a-star.edu.sg, hehe.fan.cs@gmail.com}
}

\maketitle

\begin{abstract}
 In this technical report, we present our findings from a study conducted on the EPIC-KITCHENS-100 Unsupervised Domain Adaptation task for Action Recognition. Our research focuses on the innovative application of a differentiable logic loss in the training to leverage the co-occurrence relations between verb and noun, as well as the pre-trained Large Language Models (LLMs) to generate the logic rules for the adaptation to unseen action labels. Specifically, the model's predictions are treated as the truth assignment of a co-occurrence logic formula to compute the logic loss, which measures the consistency between the predictions and the logic constraints. By using the verb-noun co-occurrence matrix generated from the dataset, we observe a moderate improvement in model performance compared to our baseline framework. To further enhance the model's adaptability to novel action labels, we experiment with rules generated using GPT-3.5, which leads to a slight decrease in performance. These findings shed light on the potential and challenges of incorporating differentiable logic and LLMs for knowledge extraction in unsupervised domain adaptation for action recognition. Our final submission (entitled `NS-LLM') achieved the first place in terms of top-1 action recognition accuracy.
\end{abstract}

\section{Introduction}
\label{sec:intro}

The EPIC-KITCHENS-100 dataset is a large-scale egocentric video dataset capturing a wide range of cooking activities. These activities, captured in various kitchens via head-mounted cameras, primarily consist of fine-grained actions involving extensive hand-object interactions. The Unsupervised Domain Adaptation (UDA) for Action Recognition Challenge aims to learn an action recognition model on a labeled source domain and generalize it to an unlabeled target domain. It draws growing interest within the community due to its potential to substantially reduce the annotation workload when deploying a trained model to datasets without annotations.

In this dataset, each action within the dataset is defined as a combination of a verb and a noun, covering 97 distinct verbs and 300 unique nouns distributed in a long-tail fashion. To improve classification efficiency, existing methods either adopt dual classification branch (one for verb and the other for noun) or train separate models for verb and noun classification. However, such approaches overlook the inherent co-occurrence relations between verbs and nouns, thereby resulting in sub-optimal performance. For instance, the module might predict an invalid action such as `rinse table', which is unlikely to occur in real-world scenarios. A recent method~\cite{Cheng2022TeamVT} proposes to apply the verb-noun co-occurrence to refine the final action predictions, leading to effective performance improvement. Despite this advancement, the application of this co-occurrence knowledge post-training means the learned features remain unaware of the co-occurrence relationships, which ultimately impedes learning efficiency. Moreover, the co-occurrence matrix, which is extracted from data annotations from the source domain, may fall short in scenarios where unseen action classes emerge in the target domain.

\begin{figure*}[t]
  \centering
  \includegraphics[width=1.0\linewidth]{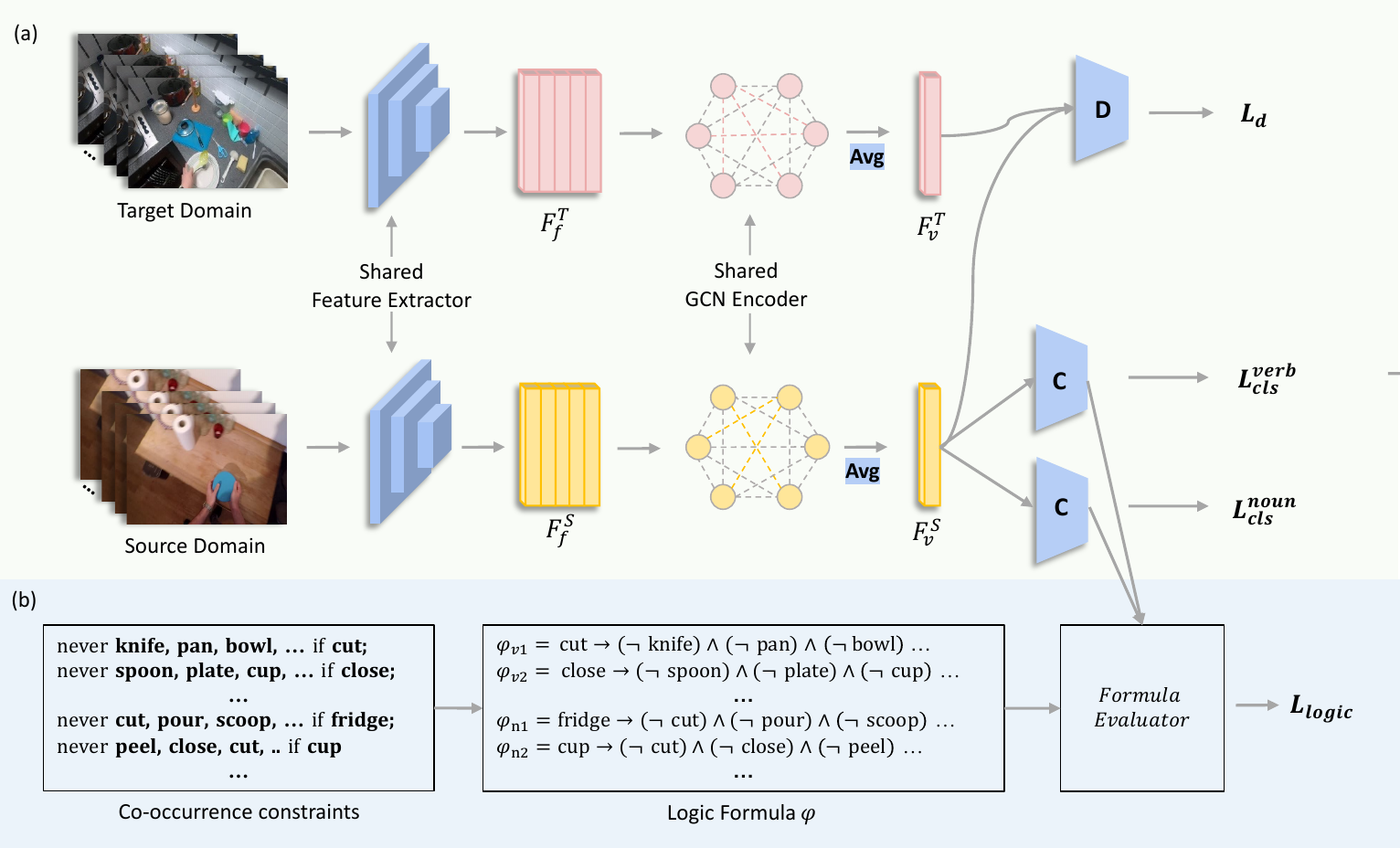}
  \caption{Overall architecture of the proposed framework. It consists of (a) the baseline framework for video domain adaptation and (2) the logic evaluation using verb-noun co-occurrence constraints. In (a), the frame-level source/target features ($\mF_f^S$/$\mF_f^T$) are extracted from video frames using a pre-trained feature extractor. Then, the extracted features are passed to a Graph Convolutional Network (GCN), followed by an average operation, to generate the video-level source/target feature ($\mF_v^S$/ $\mF_v^T$). Next, $\mF_v^S$ is passed to the action classifier, and then $\mF_v^T$ with $\mF_{vp}^S$ are aligned for video domain adaptation. $L_d$ denotes domain classification loss. $L_{\rm{cls}}^{\rm{verb}}$ and $L_{\rm{cls}}^{\rm{noun}}$ denote action classification loss for verb and noun branches, respectively. In (b), a set of constraints are generated from the co-occurrence matrix and then converted to respective logic formulas $\phi$. A formula evaluator takes both predictions and logic formulas as input and computes the logic loss $L_{\rm{cls}}$. This figure is best viewed in color.} 
\label{fig:architect}
\end{figure*}

To address these limitations, we propose a novel approach that embeds verb-noun co-occurrence relations directly into the training process by leveraging the differentiable logic loss proposed in ~\cite{xu2022don}. Specifically, we first generate a set of logic formulas from the co-occurrence matrix. Subsequently, the model's predictions are treated as the truth assignment of a co-occurrence logic formula to compute the logic loss, which measures the level of consistency between the predictions and the inherent logic constraints.
Moreover, to enhance the model's adaptability to unseen action labels, we leverage the GPT-3.5 model to generate the verb-noun co-occurrence matrix.

\section{Our Approach}
In this section, we describe the technical details of our approach. As illustrated in Fig.~\ref{fig:architect}, the overall framework consists of two components: (a) the baseline framework for video domain adaptation; (2) the logic evaluation to measure the consistency between predictions and logic constraints; and (3) the knowledge extraction from LLMs to improve model's adaptability. We will introduce each component in the following subsections.

\subsection{Baseline Framework}
To learn domain-invariant video feature representations for robust action recognition across domains, it is crucial to extract discriminative spatial and temporal relations from videos. Therefore, we design a baseline framework with three main components: (1) feature extractors to extract discriminative video features; (2) a two-branch classification module to independently predict verb and noun classes; and (3) domain adaptation.

\noindent\textbf{Feature extractors.} 
To extract discriminative feature representations from input video frames, we leverage two popular models: SlowFast~\cite{slowfast2019iccv} and MViTv2~\cite{li2021improved}. 
The SlowFast model is based on the 3D Convolutional Neural Networks, while the MViTv2 model employs a transformer-based architecture. 
Given the fact that videos may vary greatly in length, and actions performed by different subjects may occur at differing speeds, features generated by these models ($\mF_f^S$ and $\mF_f^T$) may not fully capture the temporal relations in videos. Therefore, we apply a fully-connected GCN encoder to further mine the temporal dependencies among different frames. Specifically, $\mF_f^S$ and $\mF_f^T$ are first embedded into a shared feature space via a fully-connected layer. Then the embedded features are fed into the GCN encoder. Lastly, we apply average pooling on the GCN output, generating the video-level feature representations, $\mF_v^S$ and $\mF_v^T$. This encoding process helps to generate a more robust and accurate representation of the video data, which is crucial for the subsequent video domain adaptation.

\noindent\textbf{Two-branch classifiers.} 
As described in Sec~\ref{sec:intro}, each action in the dataset is defined by a combination of a verb and a noun, with the dataset containing 97 unique verbs and 300 different nouns. These verb-noun combinations are distributed in a long-tail fashion, meaning there are many rare/invalid combinations and a few common ones. To effectively handle the long-tail distribution in action classes, we follow existing methods to design a two-branch classification module. Specifically, the module has one branch for verb class prediction and the other branch for noun class prediction. This design allows our model to specialize in each component of an action, therefore improving the classification efficiency.

\noindent\textbf{Domain adaptation.} 
To perform the alignment of features from the source and target domains, we employ a domain classifier. It aims to determine whether a given sample is from the source or target domain. To ensure effective learning and alignment, we incorporate a gradient layer between the domain classifier and our main model, inspired by TA3N~\cite{chen2019temporal}. This layer plays a crucial role in facilitating gradient back-propagation, ensuring that the updates from the domain classifier are correctly reflected in the main model. As shown in Fig.~\ref{fig:architect}, the baseline framework contains three losses: the classification losses for verb $L_{\rm{cls}}^{\rm{verb}}$, classification losses for noun $L_{\rm{cls}}^{\rm{noun}}$, and the domain classification loss $L_d$.

\subsection{Differentiable Logic Constraints}
\label{sec:logiccons}
The application of a two-branch classification module for independent verb and noun prediction neglects the inherent co-occurrence relationships between verbs and nouns, leading to sub-optimal performance. To address this limitation, we propose a novel approach that directly incorporates verb-noun co-occurrence relations into the training process. Specifically, we adapt the differentiable logic loss in~\cite{xu2022don} to account for the co-occurrence constraints between verbs and nouns in action recognition.

As illustrated in Fig.~\ref{fig:architect} (b), the first step in our approach involves generating a set of co-occurrence constraints from the co-occurrence matrix. This matrix is created from the training data, capturing the observed relations between verbs and nouns. In this work, we assume that verb and noun combinations that do not occur in the training dataset are invalid. Then, these constraints are converted into respective logic formulas. Following the generation of these logic formulas, we treat the model’s predictions as the truth assignment of a co-occurrence logic formula. In other words, each prediction made by the model is evaluated in the context of the logic formulas derived from the co-occurrence matrix. This evaluation results in the computation of the logic loss, which measures the consistency between the predictions and the logic constraints.

By minimizing the logic loss during training, the model's predictions are encouraged to satisfy the logic constraints generated from the data. This process allows the model to leverage the inherent co-occurrence relations between verb and noun classes by providing explicit supervision signals, leading to improved performance for action recognition.

\subsection{Knowledge Extraction from LLMs}
\label{sec:gpt}
As discussed Sec~\ref{sec:logiccons}, the co-occurrence matrix derived from the training data plays a critical role in our method. However, the matrix generated from training data may overlook potential verb-noun combinations that were not present in the training set but could occur in real-world situations. This could potentially limit the model's adaptability to unseen action classes.

To overcome this limitation, we propose an innovative approach to generate the co-occurrence matrix using GPT3.5, OpenAI's advanced language model. This approach allows us to leverage the world knowledge embedded in LLMs to predict more complete verb-noun co-occurrences. Specifically, we query GPT3.5 API with each combination of verb and noun by asking whether the action makes sense in the context of daily cooking activities. Then the results are used to construct a new co-occurrence matrix, which is used in the same manner as the one generated from training data.

By integrating the language understanding capabilities of GPT3.5, we expect our model to gain a more comprehensive understanding of verb-noun relations, extending beyond that observed in the training data. In essence, this approach allows our model to `imagine' new action classes, thereby increasing its adaptability to unseen scenarios.

\section{Experiments}
\label{sec:exp}

\begin{figure*}[t]
  \centering
  \includegraphics[width=1.0\linewidth]{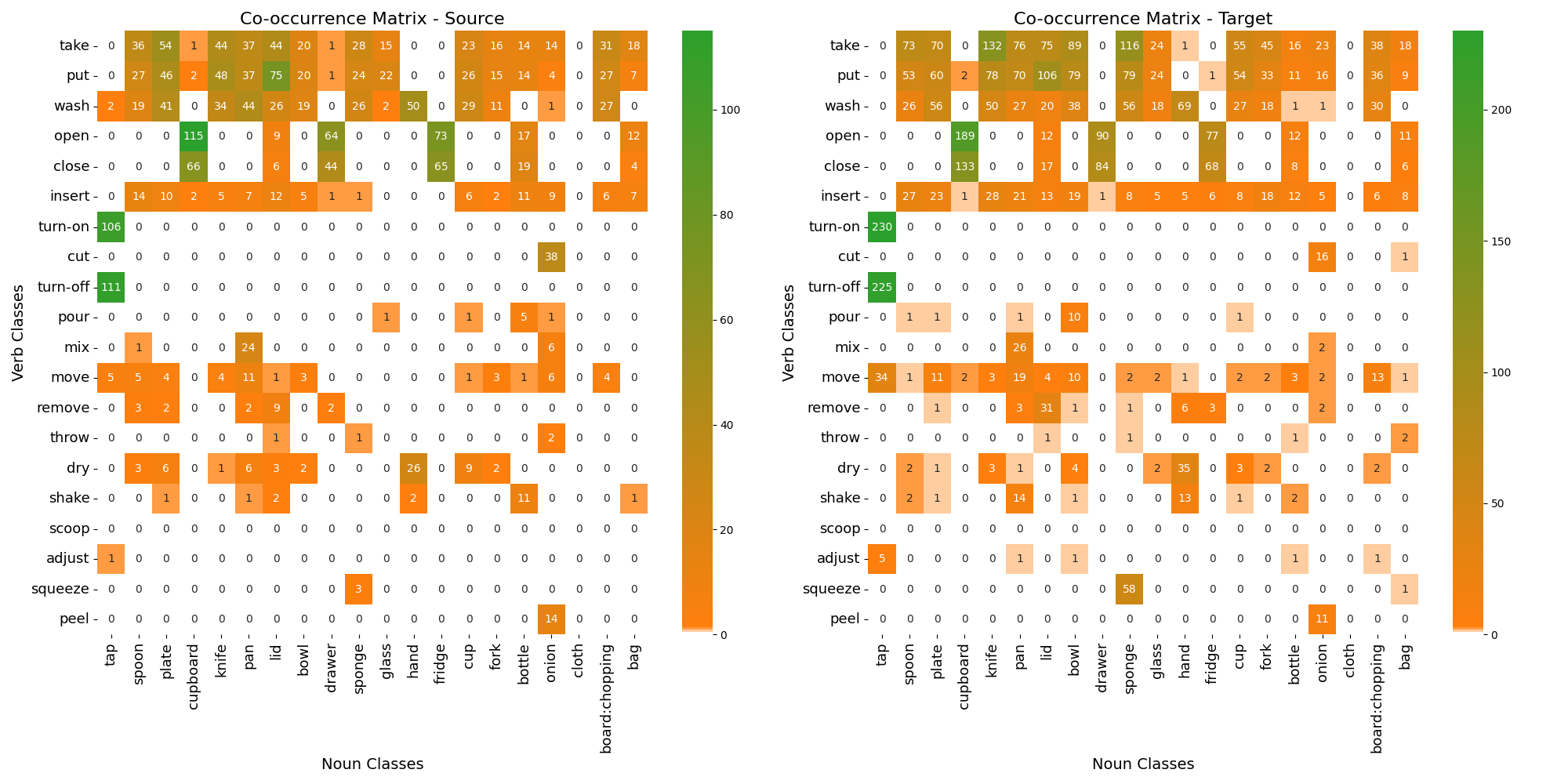}
  \caption{Visualization of verb-noun co-occurrence matrix, generated from source and target domain of the validation set. We select a sample of 20 classes from both verbs and nouns for display. This figure is best viewed in color.} 
\label{fig:matrix}
\end{figure*}

\subsection{Implementation Details}
We train three variants of the SlowFast~\cite{slowfast2019iccv}, including SlowFast with ResNet50, SlowFast with ResNet101, and SlowOnly (using only the slow path in SlowFast) with ResNet50. For each of the three variants, the model is trained for 60 epochs using synchronized SGD training as in~\cite{slowfast2019iccv}. The input number of frames is set as $32$ and $8$ for the fast and slow paths, respectively. The batch size is set as $64$. We also train MViTv2 for 100 epochs using Adam optimizer, and the input number of frames is set as $32$. During feature extraction, we extract the features from the last convolutional layer and apply average pooling to generate the frame-level feature representations. The feature dimension for SlowOnly-ResNet50 is $2048$, and the feature dimensions for SlowFast-ResNet50 and SlowFast-ResNet101 are $2304$, and the feature dimensions for MViTv2 is $768$. 

Following the challenge guidelines, our full model is first trained on the validation set for algorithm validation and hyper-parameters tuning and then retrained on the training set with selected hyper-parameters. 
During training, the parameters of the feature extractors are fixed, while the other parameters are learned using an initial learning rate at $3 \times 10^{-3}$. The training process spans 30 epochs, with the learning rate experiencing a decimation by a factor of 10 at the 10th and 20th epochs. 
In our final model, we use the verb-noun co-occurrence matrix generated from the dataset for training. 

\begin{table}[ht]
\caption{The comparison of models trained with different loss functions on the EPIC-KITCHENS-100 validation set. The `Base' model refers to the one trained solely using action classification loss and domain classification loss. `Base+LR' indicates the model that incorporates the logic loss using the co-occurrence matrix generated from the dataset. `Base+LR+GPT3.5' indicates the model that incorporates the logic loss using the co-occurrence matrix generated by GPT3.5. The best performance is marked in bold.}
\centering
\scalebox{0.75}{
\begin{tabular}{c|c|c|c|c|c|c}
\hline
\multirow{2}{*}{Method} & \multicolumn{3}{c|}{Top-1 Accuracy (\%)} & \multicolumn{3}{c}{Top-5 Accuracy (\%)} \\ \cline{2-7} 
 & Verb & Noun & Action & Verb & Noun & Action \\ \hline
Base & 52.36  &  34.83 & 23.83 &80.89  & 58.68  & 51.69 \\ \hline
Base+LR & \textbf{52.75} & 35.02 & \textbf{24.36} & \textbf{81.33} & \textbf{59.12} & 52.57 \\  \hline
Base+LR+GPT3.5& 52.65 & \textbf{35.22} & 24.16 & 81.19 & 58.93 & \textbf{52.71} \\ \hline

\end{tabular}
}
\label{tab:ablation}
\end{table}

\subsection{Results and Analysis}
Table~\ref{tab:ablation} shows the model performance on the target validation set, using the RGB and Flow features extracted from the pre-trained SlowOnly-ResNet50 model. We observe a modest performance improvement of 0.5\% in terms of top-1 action accuracy on the validation set when incorporating the verb-noun co-occurrence relation into model training. This improvement is not as significant as expected. There are mainly two reasons. Firstly, the co-occurrence matrix is generated from action labels in the source domain, and it is possible that new, unseen action classes may emerge in the target domain, as shown in Fig.~\ref{fig:matrix}. This may hinder the model's ability to accurately classify such actions. Secondly, the co-occurrence constraints may not impose strong enough constraints on the model's predictions. 
More robust supervision could be introduced through spatial relation constraints, which specify certain relations between hand and object, and thereby provide stronger guidance for the model's learning. This could enhance the model's ability to recognize actions.

\begin{table}[t]
\caption{Comparison of our final model with others on the EPIC-KITCHENS-100 test set.`Baseline' denotes the baseline results provided by the challenge organizer,
and `Ours' denotes our final model after the model ensemble.}
\centering
\scalebox{0.85}{
\begin{tabular}{c|c|c|c|c|c|c}
\hline
\multirow{2}{*}{Method} & \multicolumn{3}{c|}{Top-1 Accuracy (\%)} & \multicolumn{3}{c}{Top-5 Accuracy (\%)} \\ \cline{2-7} 
 & Verb & Noun & Action & Verb & Noun & Action \\ \hline
Baseline & 46.91  &  27.69 & 18.95 &72.70  & 50.72  & 30.53 \\
Ours & \textbf{58.22}  &  \textbf{40.33} & \textbf{30.14} &\textbf{84.26}  & \textbf{64.71}  & \textbf{49.18} \\ \hline
\end{tabular}
}
\label{tab:test}
\end{table}

\noindent\textbf{Can LLMs help improve model performance?} 
As discussed in Sec~\ref{sec:gpt}, we propose to employ GPT-3.5 to generate a more general co-occurrence matrix. Table~\ref{tab:ablation} shows the effect of using a GPT-generated matrix, which leads to a 0.2\% drop in action recognition accuracy, but a 0.2\% rise in top-1 noun accuracy. 
This phenomenon can be attributed to multiple potential factors. First, the weakening of regularization might have contributed to the change. Based on our observation, the matrix generated by GPT-3.5 contains more valid action classes. Although a broader perspective can make the model more adaptable to new action classes, it may also lead to looser supervision to guide model training. Second, both data-derived and GPT-generated co-occurrence matrices contain some noise. We observe that there are some invalid actions existing in the dataset annotations (e.g., `put fridge' shown on the right of Fig~\ref{fig:matrix}), while the GPT-generated matrix is very sensitive to prompts and may not be perfect. 

\subsection{Model Ensemble}
To leverage the complementary nature of predictions from multiple models, we adopt an ensemble approach to combine the predictions from various models.
These models are trained on features extracted using the three variants of the SlowFast~\cite{slowfast2019iccv} and MViTv2 models. In addition, we incorporate the predictions from AC-VDA~\cite{Cheng2022TeamVT}, which introduces an action-aware domain adaptation framework for this task. Using the aggregation strategy in~\cite{huang2021towards}, we aggregate the action probabilities from each model to formulate our final predictions. Our combined results, shown in Table~\ref{tab:test}, have secured the first position for the top-1 action accuracy in the EPIC-KITCHENS-100 UDA Challenge for Action Recognition 2023.

\section{Conclusion}
In conclusion, this report has presented our study on differentiable logic and LLMs for EPIC-KITCHENS-100 Unsupervised Domain Adaptation Challenge for Action Recognition 2023. Our method leverages a differentiable logic loss during training to incorporate the co-occurrence relations between verbs and nouns. To enhance the model's adaptability, we experimented with using OpenAI's advanced language model, GPT-3.5, to generate the verb-noun co-occurrence matrix. Although the performance does not fully meet our expectations, it represented a pioneering effort to incorporate the knowledge from LLMs in the task of UDA for action recognition. With further performance increase from the model ensemble, our final submission achieves the first rank on the leaderboard in terms of top-1 action recognition accuracy.

\section*{Acknowledgments}
We would like to thank Dr. Joo Hwee Lim for his continuous support and useful guidance. This research is supported by the Agency for Science, Technology and Research (A*STAR) under its AME Programmatic Funding Scheme (Project \#A18A2b0046). 

{\small
\bibliographystyle{ieee_fullname}
\bibliography{egbib}

\begin{thebibliography}{1}\itemsep=-1pt

\bibitem{chen2019temporal}
Min-Hung Chen, Zsolt Kira, Ghassan AlRegib, Jaekwon Yoo, Ruxin Chen, and Jian
  Zheng.
\newblock Temporal attentive alignment for large-scale video domain adaptation.
\newblock In {\em Proceedings of the IEEE/CVF International Conference on
  Computer Vision}, pages 6321--6330, 2019.

\bibitem{Cheng2022TeamVT}
Yi Cheng, Dongyun Lin, Fen Fang, Hao~Xuan Woon, Qianli Xu, and Ying Sun.
\newblock Team {VI-I2R} technical report on epic-kitchens-100 unsupervised
  domain adaptation challenge for action recognition 2022.
\newblock 2023.

\bibitem{slowfast2019iccv}
Christoph Feichtenhofer, Haoqi Fan, Jitendra Malik, and Kaiming He.
\newblock Slowfast networks for video recognition.
\newblock pages 6201--6210, 10 2019.

\bibitem{huang2021towards}
Ziyuan Huang, Zhiwu Qing, Xiang Wang, Yutong Feng, Shiwei Zhang, Jianwen Jiang,
  Zhurong Xia, Mingqian Tang, Nong Sang, and Marcelo~H Ang~Jr.
\newblock Towards training stronger video vision transformers for
  epic-kitchens-100 action recognition.
\newblock {\em arXiv preprint arXiv:2106.05058}, 2021.

\bibitem{li2021improved}
Yanghao Li, Chao-Yuan Wu, Haoqi Fan, Karttikeya Mangalam, Bo Xiong, Jitendra
  Malik, and Christoph Feichtenhofer.
\newblock Improved multiscale vision transformers for classification and
  detection.
\newblock {\em arXiv preprint arXiv:2112.01526}, 2021.

\bibitem{xu2022don}
Ziwei Xu, Yogesh Rawat, Yongkang Wong, Mohan~S Kankanhalli, and Mubarak Shah.
\newblock Don't pour cereal into coffee: Differentiable temporal logic for
  temporal action segmentation.
\newblock {\em Advances in Neural Information Processing Systems},
  35:14890--14903, 2022.

\end{thebibliography}
}

\end{document}